\documentclass[11pt]{article} 

\usepackage[utf8]{inputenc}
\usepackage[T1]{fontenc}
\usepackage{textcomp}
\usepackage{newunicodechar}
\newunicodechar{−}{\textminus}
\usepackage{amsmath,amssymb}
\usepackage{booktabs}
\usepackage{gensymb}
\usepackage{url,hyperref,microtype,subcaption}
\usepackage[onehalfspacing]{setspace}
\usepackage{biblatex}
\usepackage[section]{placeins}
\usepackage{graphicx}   
\usepackage[margin=1in]{geometry}

\title{Neural-Network Maxent: a general extension with learned nonlinearity, applied to time-series for Desert Locust distribution modelling}

\author{%
Alessandro Grassi$^{1,*}$, Edoardo Kimani Bellotto$^{1}$, Wassim El Azami$^{2}$, Sabrina Outmani$^{1}$, \\ and Maximilien Houel$^{2}$\\[6pt]
\small $^{1}$SISTEMA GmbH, Vienna, Austria\\
\small $^{2}$Beyond EO, Paris, France\\[6pt]
\small $^{*}$Corresponding author: Alessandro Grassi (\texttt{grassi@sistema.at})
}
\date{}

\begin{document}
\maketitle

\begin{abstract}Species Distribution Modelling (SDM) is essential for understanding how environmental conditions shape biodiversity, particularly for destructive pests such as the Desert Locust (\textit{Schistocerca gregaria}), whose breeding dynamics are tightly coupled to rapidly evolving environmental conditions. Maxent has become the dominant method for presence-only data, but its reliance on a linear combination of hand-chosen feature transforms limits its ability to capture the nonlinear, temporal relationships common in ecological monitoring, where covariates such as precipitation, soil moisture, and vegetation indices evolve meaningfully over time. Standard implementations flatten time-series covariates into independent features, discarding sequential structure that carries critical signal. We introduce RNN-Maxent, an extension of the Maxent framework that replaces the fixed feature dictionary with a neural network---specifically a Gated Recurrent Unit (GRU)---trained end-to-end via backpropagation. The approach preserves Maxent's presence-only statistical foundations, background normalization, and probability calibration, differing only in that the nonlinearity is learned from data rather than fixed in advance. We apply RNN-Maxent to map suitable habitat for the Desert Locust using 50-day environmental time series derived from ERA5-Land, MODIS, and Sentinel-3, maintaining a 7-day gap between covariates and presence records to yield forecasting behaviour. Compared against standard Maxent, RNN-Maxent improves performance across metrics (ROC-AUC $0.862\pm0.036$ vs. $0.792$; F1 $0.671\pm0.056$ vs. $0.590$). Critically, RNN-Maxent flags a smaller fraction of the samples while capturing similar amount of Desert Locust findings, producing a tighter, more actionable footprint for control operations. Building on these suitability predictions, we further propose a stochastic migration model --- a biased random walk on a 2D lattice --- that ingests RNN-Maxent outputs and integrates Destination Earth Digital Twin data to simulate adult locust dispersion and swarm formation, offering a complementary tool for early warning systems operating over large, remote, and hard-to-access regions within the processing domain, from West Africa to Northwest Asia.
\end{abstract}


\section{Introduction}

Understanding how environmental conditions affect geographical distribution of species is extremely important for studying biodiversity and estimating how it might vary in response to climate change or weather conditions. Some species, of both animals and plants, become extremely dangerous when growing in large numbers, potentially damaging the environment around them. The Desert Locust (\textit{Schistocerca gregaria}) makes a concrete example of such a threatening species. Understanding their behavior would grant better precision in control operations aiming at containing such pests.

Among Species Distribution Modelling (SDM) approaches, Maxent \cite{phillips2006maxent} has established itself as the dominant method for presence-only data \cite{Almeida2025Ticks} \cite{OzkanTumer2025SDM}{.} However, its reliance on a linear combination of input features limits its capacity to capture complex, nonlinear relationships between environmental variables and species presence. This limitation becomes particularly evident when the input data carries a temporal structure, as is common in ecological monitoring scenarios where covariates such as precipitation, soil moisture, and vegetation indices evolve over time in ways that are ecologically meaningful. Standard Maxent implementations typically flatten time-series covariates into independent features, discarding the sequential ordering and temporal dependencies that carry critical signal.

This paper proposes an extension of the Maxent framework, creating a model that keeps the qualities of the original approach while adding a neural network at its core, capable of understanding non-linear patterns. The specific implementation of this novel approach discussed in this article uses a recurrent neural network, specifically a Gated Recurrent Unit (GRU)~\cite{Chung2014GRU}{,} in place of the linear feature transformation; we refer to it as RNN-Maxent. RNN-Maxent preserves the presence-only statistical foundations and probability-of-presence output while gaining the capacity to learn nonlinear temporal dynamics from environmental time series, trained end-to-end via backpropagation~\cite{rumelhart1986backpropagation}{.} To avoid confusion, from now on we refer to the original Maxent as Vanilla-Maxent.

To demonstrate its effectiveness, we apply RNN-Maxent to the task of mapping suitable habitat for the Desert Locust, which is considered one of the most destructive migratory pests on Earth and a species whose breeding dynamics is tightly coupled to rapidly evolving environmental conditions \cite{symmons2001guidelines}{.} Results are compared directly against standard Maxent, validating the advantage of explicitly modeling temporal covariate dynamics in presence-only species distribution modeling.

\section{State of the art}

Species Distribution Modeling (SDM) has provided essential insights into how environmental variables govern the behaviour of biodiversity. Despite the abundance of machine learning (ML) algorithms, SDM rarely deals with standard ML scenarios, as most of the available datasets are fully unbalanced, where only occurrence data is available, consisting of a positive-only dataset \cite{ward2009presenceonly}{.}

Given this setting, ad-hoc frameworks have been developed to address this limitation, which can be broadly categorized into:
\begin{itemize}
    \item the generation of pseudo-absences, treated as negative samples within standard ML algorithms;
    \item the use of presence-only (positive-only) modeling approaches.
\end{itemize}

The contribution of this work falls within the second category, extending the capabilities of a widely used model: Maxent \cite{phillips2006maxent}{.}

Let \(z \in \mathbb{R}^d\) denote the environmental covariates, and
\(y \in \{0,1\}\) the presence indicator, we can define \(f(z)\) as the probability density of the covariates over all the study area, \(f_1(z)\) as the probability density of the covariates at the finding locations, \(y=0\) as the absence of species, and \(y=1\) as the presence of species. The quantity that Maxent wishes to estimate is \(Pr(y=1 | z)\), following the Bayes' rule: \(Pr(y=1 | z) = \frac{f_1(z)Pr(y=1)}{f(z)}\). The only element missing in the latter formula is \(Pr(y=1)\), known as prevalence, which will be estimated based on certain assumptions that, along with the concept of sampling bias, is not in the scope of this chapter. More details can be found in the original paper. The ratio \(f_1(z)/f(z)\) is known as the raw output, giving insights on feature importance and estimates the relative suitability of one place vs another.

In order to estimate \(f(z)\), Maxent draws random points across the area of interest, known as background points, \(f_1(z)\) is estimated as the distribution that fits the environmental variables at the presence location and that minimizes the cross-entropy between \(f_1(z)\) and \(f(z)\), i.e., maximizing the entropy of \(f_1(z)\). As explained in \cite{phillips2006maxent}{,} minimizing the relative entropy results in a Gibbs distribution, formulated as: \(f_1(z)=f(z)e^{\eta(z)}\) where \(\eta(z)=b+w z\), where \(b\) and \(w\) are the parameters that have to be fitted. This formulation reveals that the log-density ratio can be expressed as a linear function of the covariates, establishing a connection with generalized linear models (i.e., logistic regression):

\[
    \frac{f_1(z)}{f(z)} = e^{b+w z} \Rightarrow log(\frac{f_1(z)}{f(z)}) = b+w z  \quad
\] \cite{elith2010explanation_maxent}

As described by \cite{fithian2014poisson}{,} the parameters can be learned by fitting a logistic regression with a weight on negative samples that tends to infinity, where positive samples correspond to presence location, and negative samples are background locations. Practically high weight to negative samples, or low weight to positive samples are sufficient.

Estimating the coefficients $b$ and $w$ allows to identify the shape of the density ratio $f_1(z)/f(z)$, but not its absolute scale: the model still needs a normalising constant before it can be evaluated as a density and converted to a probability of presence. This constant, denoted $\alpha$, is the offset that makes the exponential model $\exp(\eta(z) + \alpha)$ sum to one over the background sample, and is computed in closed form as $\alpha = -\log \sum_{z \in \mathcal{B}} \exp(\eta(z))$, where $\mathcal{B}$ is the set of background points\cite{fithian2014poisson}{.} Alongside $\alpha$, Maxent also computes the entropy $H$ of the fitted distribution over the background, $H = -\sum_{z \in \mathcal{B}} p(z) \log p(z)$ with $p(z) = \exp(\eta(z) + \alpha)$.
Both $\alpha$ and \(H\) are derived post-hoc, in a single pass over the background sample after fitting, and are not learned by the regression itself. Their primary use is to convert the raw linear predictor \(\eta(z)\) into a calibrated probability of presence.

\section{Methodologies}
In this section, we detail the mathematical formulation and implementation of the proposed approach that combines neural network architectures with the Maxent framework to model presence‑only species distribution data. We will focus on our use case aimed at mapping the suitable Desert Locust (DL) ecosystem by analyzing time-series data through a recurrent neural network (RNN). Following the previous section, we first formalize the statistical concepts of our model and the assumptions that guide its implementation. Finally, we outline the implementation workflow and training procedures that support reproducibility and empirical evaluation.

\subsection{Formal definition}
\begin{figure*}[ht]
    \centering
    \includegraphics[width=1\linewidth]{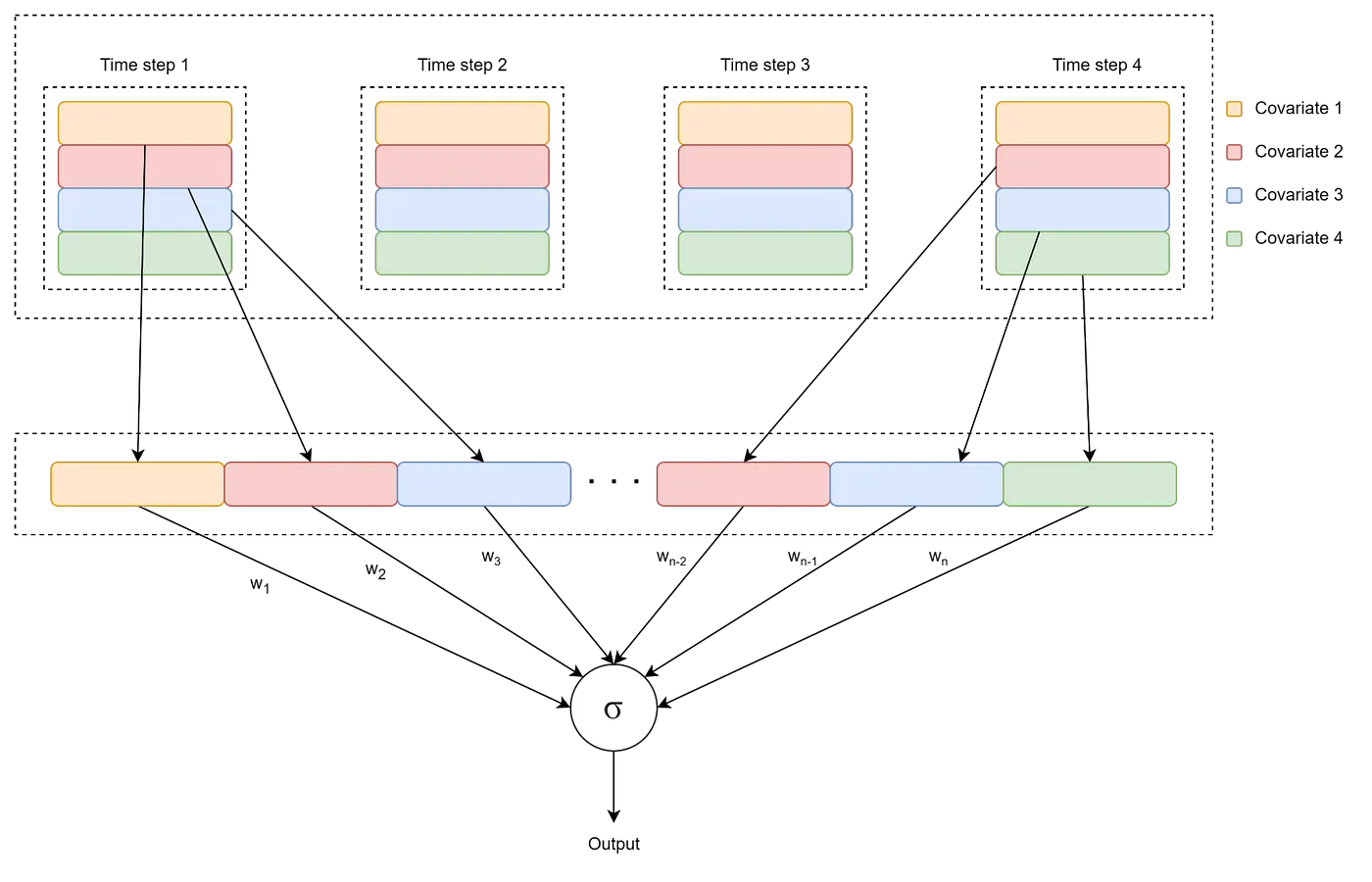}
    \caption{Flattening of input time series for Vanilla Maxent. The time series gets flattened, transforming each variable at each time step as independent from the others.}
    \label{fig:flattened_maxent}
\end{figure*}

As we explain in the State of the Art, our objective is to find the ratio between the density function of the environmental variables at the locations of occurrence \(f_1(z)\) and the density function of the environmental variables throughout the area of interest \(f(z)\), where \(f_1(z)\) minimizes the cross-entropy with \(f(z)\). As described by \cite{elith2010explanation_maxent} and \cite{fithian2014poisson}{,} the solution of the problem can be simplified to a logistic regression problem where the background sample has a theoretically infinite and practically high weight. In this formulation the nonlinearity in \(\eta\) is supplied a priori, through a fixed dictionary of hand-chosen feature transforms (i.e., quadratic, step function). Furthermore, when data have a temporal component, it has to be flattened to produce a viable input for the regression, as in Figure \ref{fig:flattened_maxent}, losing temporal information.

\begin{figure*}[ht]
    \centering
    \includegraphics[width=1\linewidth]{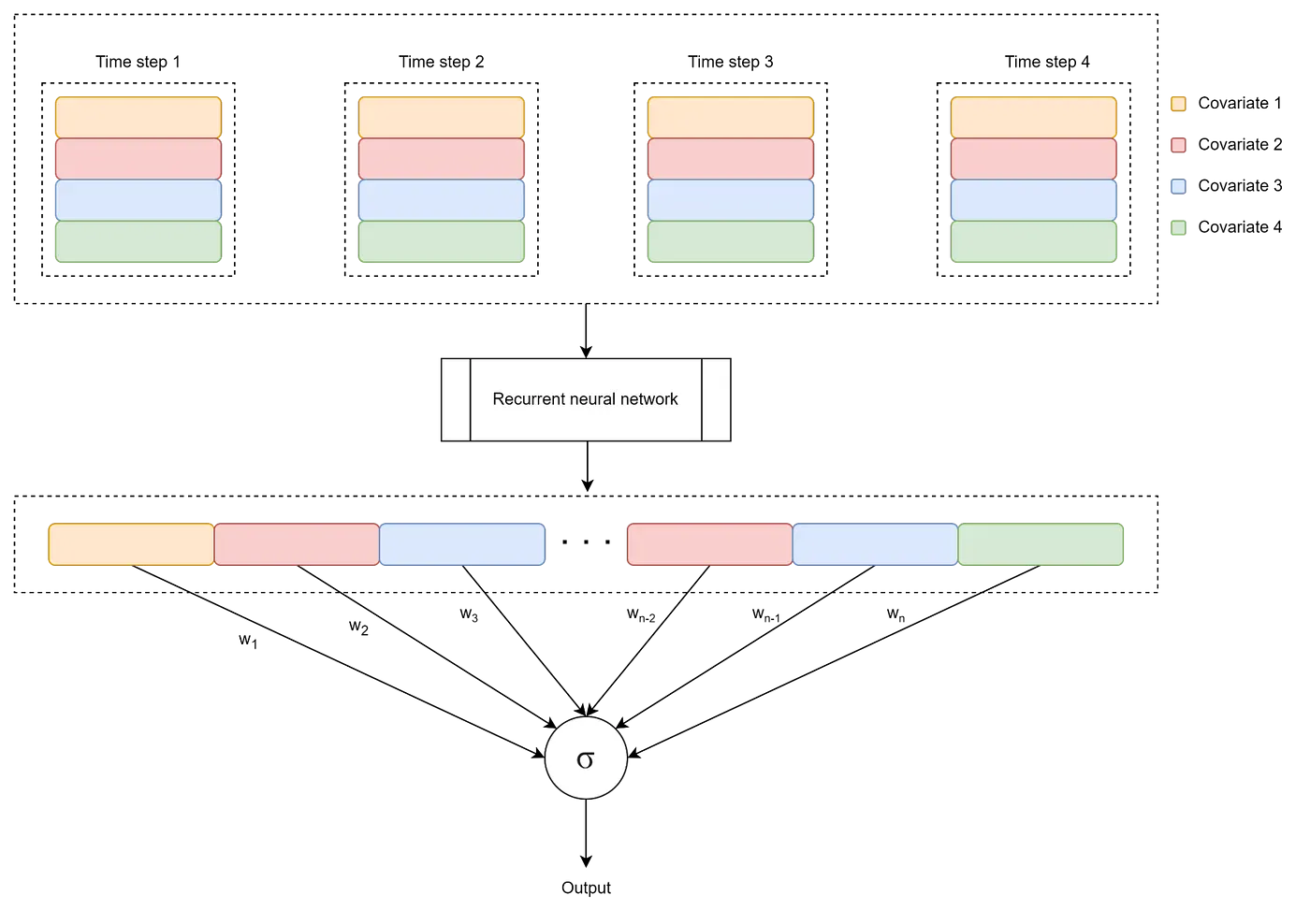}
    \caption{With RNN-Maxent, the input time series is fed to a recurrent neural network, which produces a hidden representation then used by the Maxent module. During training the hidden representation is passed through a logistic operator.}
    \label{fig:rnn_maxent}
\end{figure*}

As regression parameters are optimized through a form of gradient descent, any derivable transformation before the last linear layer can be applied and optimized by backpropagation \cite{rumelhart1986backpropagation}{.} We start by replacing this fixed feature dictionary with a neural network, feeding the original data to it and taking its last hidden layer as the input to the linear Maxent readout, as described by this formula:
\[
    \eta(z) = b + w\phi(z; \theta)
\]
where \(\phi\) is a neural network with parameters \(\theta\) learnt jointly with \(b\) and \(w\). This can be observed in Figure \ref{fig:rnn_maxent}.
Once the training is complete, the model is fed with all the negative samples so to compute the \(\alpha\) and \(H\) parameters that will be used to properly compute the probability of presence as in the original Maxent (see State of the art for deeper understanding).

Conditioned on the learned feature map \(\phi(z; \theta)\), neural-network-based Maxent is identical to the original version: the linear readout, the background normalization \(\alpha\), the entropy, and the probability calibration are all unchanged. What differs is solely the origin of the nonlinearity: the feature map is learned from the data rather than fixed in advance, allowing the model to fit nonlinear relationships that a hand-chosen dictionary may neglect.

\subsection{Desert Locust suitable ecosystem estimation}
In the following section we'll explain how we've been implementing neural-network based Maxent to map the suitable ecosystem for the Desert Locusts, using time-series data to model the complex relationship between temporal evolving data and Desert Locust presence. Our implementation, from hereafter referred to as RNN-Maxent, chooses a recurrent neural network (RNN) to fit the covariates.

\subsubsection{Dataset}

Desert Locust presence data used for training, validation and testing has been created and maintained by the Locust Watch of the FAO (Food and Agriculture Organization of the United Nations). This dataset is divided in four subsets each referring to a different development stage of the locusts finding reported, the ones used for this study are: Hoppers and Bands, both referring to young wingless locusts. Focusing on wingless samples is fundamental because they move slowly, allowing to use longer periods of historical data on the location they were found \cite{FAOLocustHub}{.}

Raw database records frequently contain multiple reports at the same location on the same day arising from different survey teams or data-entry artefacts. To prevent such duplicates from artificially inflating local density estimates, a deduplication step was applied before any further processing. Coordinates were snapped to a regular grid with a spacing of $0.05$° in both latitude and longitude, and each unique triplet (snapped longitude, snapped latitude, calendar date) was retained at most once.

The information used to correlate Desert Locust presence is environmental data downloaded from ERA5-Land \cite{MunozSabater2021ERA5Land}{,} MODIS \cite{justice1998modis}{,} and Sentinel-3 \cite{Donlon2012Sentinel3}{.} A 7 days gap is kept between the covariates and locust presence ground truth, so to obtain a forecasting behaviour. For each locust finding is associated a 50 days time series where each time step is a 5-days average of the following variables:
\begin{itemize}
    \item Soil water content layer 1 (ERA5-Land): knowing the amount of water in the soil is important since Desert Locusts need to lay eggs in moist soil;
    \item NDVI (MODIS and Sentinel-3): the Normalized Difference Vegetation Index provides a measure on how healthy and dense the vegetation is. Due to MODIS decommisioning, data from MXD13C1 \cite{Didan2021MOD13C1} \cite{Didan2021MYD13C1} was used only during training, while for testing NDVI data from Sentinel-3 SY\_2\_V10 was used \cite{CopernicusSY2V10}{.}
    \item Total precipitation (ERA5-Land): this variable is important because weather conditions affect the development of Desert Locusts.
    \item Temperature (ERA5-Land): this variable correlates with the development rate of both egg and born individuals.
\end{itemize}

\subsubsection{Implementation details}
RNN-Maxent has been implemented in python with PyTorch for model definition and PyTorch-lightning for reproducibility. The neural network we chose is a Gated Recurrent Unit (GRU) \cite{gru2014original}{.}

The dataset has been split into 3 subsets by time, samples within years 2000 and 2018 were used for training, samples in year 2019 were used for validation and hyperparameter selection, and samples in years 2020 and 2021 were used for testing.

Vanilla-Maxent is a convex optimisation problem, meaning it has a single global optimum. As a result, gradient descent will always converge to the same solution regardless of initialisation. This does not hold for RNN-Maxent: because the core function is a neural network, the problem is no longer convex, and mini-batch training introduces additional stochasticity. Consequently, different runs with different initialisations can yield different results, potentially converging to distinct local optima. Due to this issue, the RNN-Maxent model has been trained in 10 different runs each with different seeds, affecting parameters initialization and sample ordering. The scores that will be shown in the Results section will show the mean, standard deviation, and range over the 10 runs.

\section{Results}
In this section, we present the empirical evaluation of the proposed RNN-Maxent framework for modeling DL habitat suitability. We compare RNN-Maxent and Vanilla-Maxent on the same test dataset, proving the effectiveness of the proposed approach. The test dataset is made up of DL findings from years 2020 and 2021 from all over the FAO study region. It contains a total of $75775$ samples, $22233$ of which real DL findings and $53542$ are background points (i.e., randomly drawn across the area of interest).

\subsection{Discrimination}
\begin{figure*}[ht]
    \centering
    \includegraphics[width=1\linewidth]{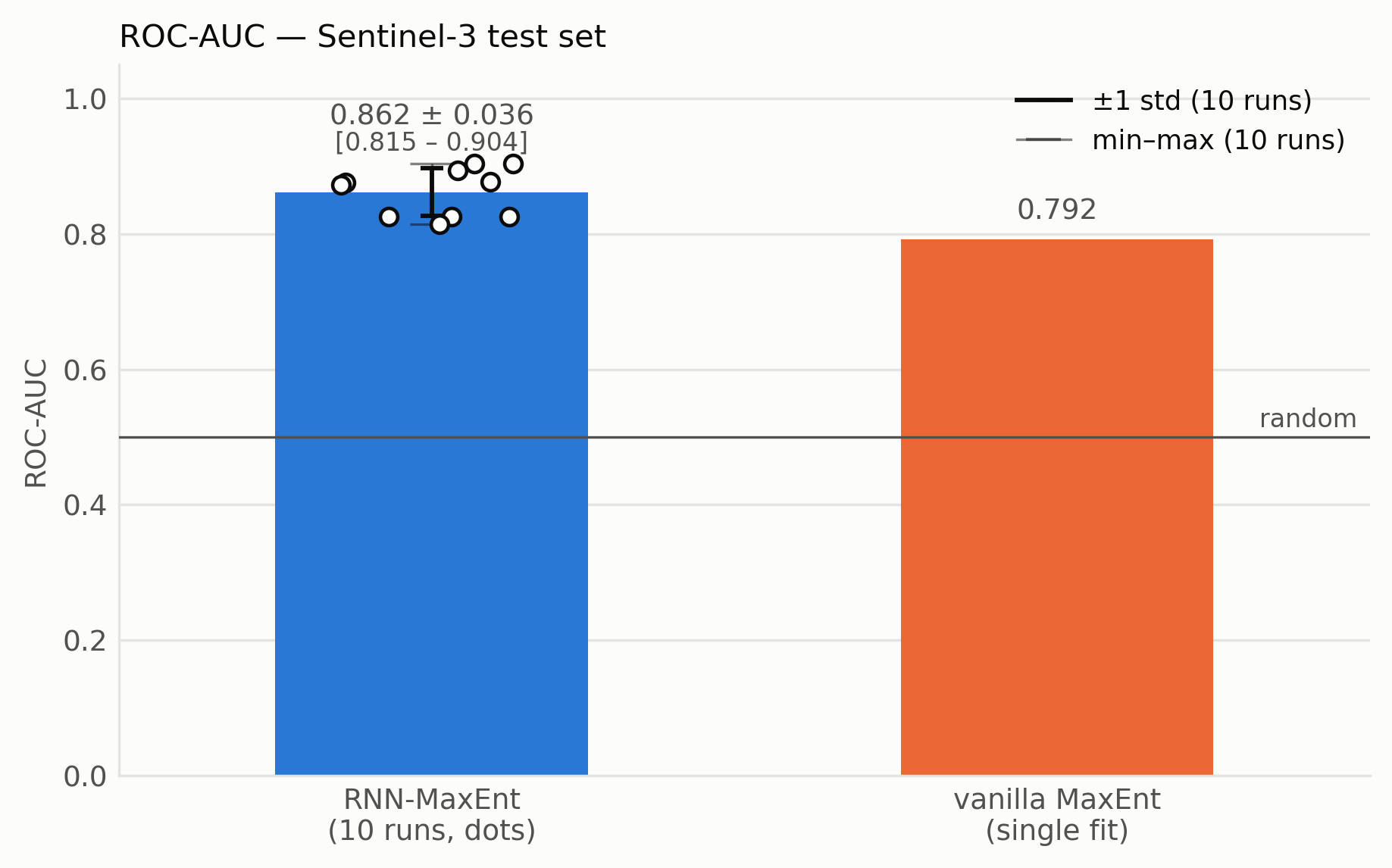}
    \caption{RNN-Maxent improves threshold-independent discrimination over Vanilla Maxent on the Sentinel-3 test set (n = 75,775). Bars give the mean ROC-AUC over 10 independently seeded RNN-Maxent runs (blue) and the single deterministic Vanilla Maxent fit (orange). Open circles are the 10 individual runs, jittered horizontally for readability. The thick whisker spans ±1 standard deviation across runs and the thin wide-capped whisker the observed minimum–maximum; the baseline is a single fit and so carries neither. The horizontal rule at 0.5 marks chance performance. RNN-Maxent reaches 0.862 ± 0.036 (range 0.815–0.904) against 0.792 for Vanilla Maxent, a gain of +0.070 in the mean, and all 10 runs — including the weakest — exceed the baseline. Bars are drawn from zero to avoid exaggerating the difference.}
    \label{fig:roc_auc_score}
\end{figure*}

\begin{figure*}[ht]
    \centering
    \includegraphics[width=1\linewidth]{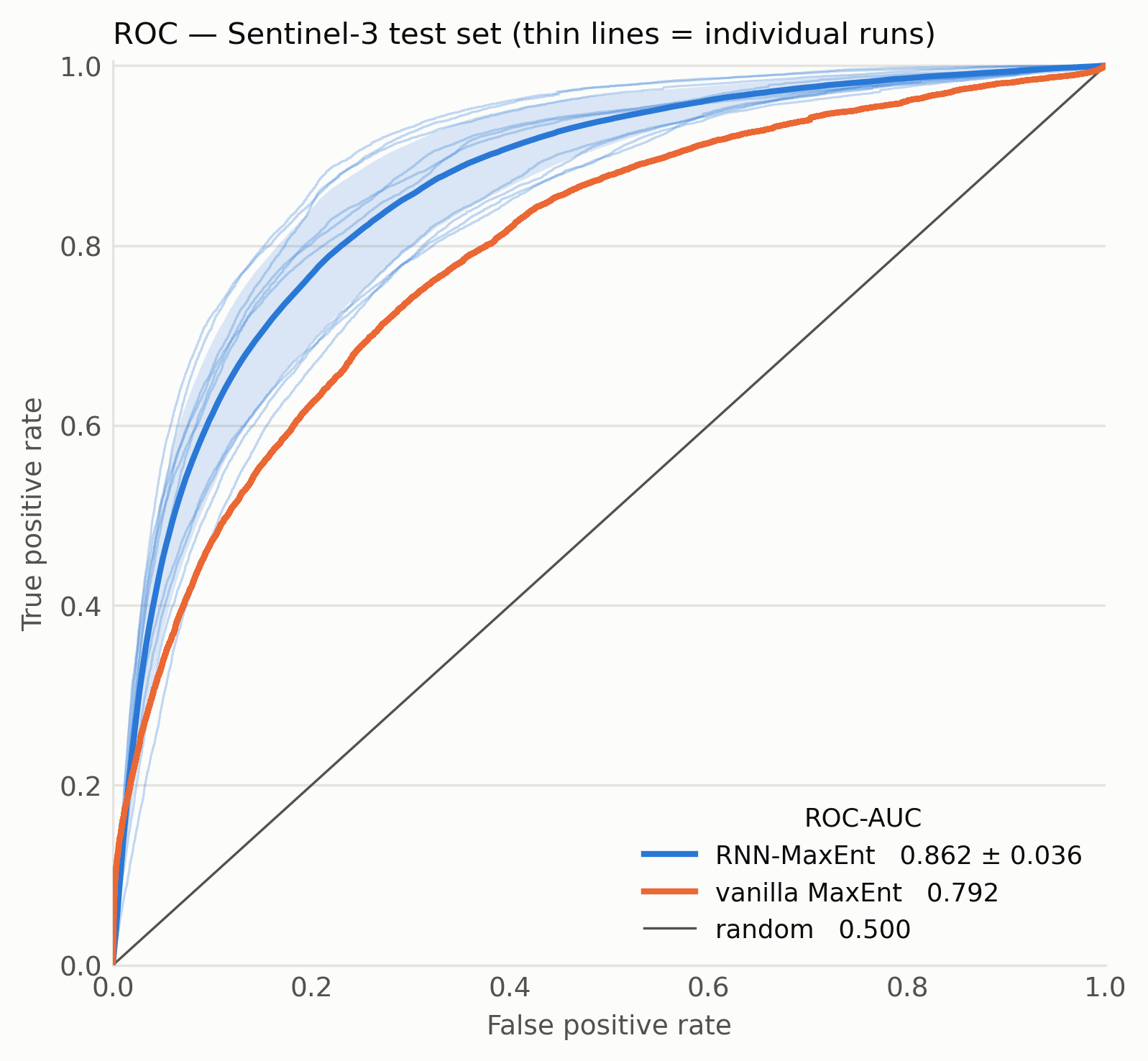}
    \caption{RNN-Maxent dominates the Vanilla Maxent ROC curve across essentially the whole operating range. Receiver operating characteristic curves on the Sentinel-3 test set (n = 75,775). Thin blue lines are the 10 individually seeded RNN-Maxent runs; the heavy blue line is their mean true-positive rate, interpolated onto a common false-positive-rate grid, and the surrounding band spans ±1 standard deviation across runs. The orange line is the single deterministic Vanilla Maxent fit and the diagonal marks chance. Areas under the curve are 0.862 ± 0.036 for RNN-Maxent and 0.792 for Vanilla Maxent.}
    \label{fig:roc_auc_curves}
\end{figure*}

RNN–Maxent discriminated presence from background substantially better than the Vanilla Maxent baseline. Averaged over the ten runs, ROC-AUC was 0.862 ± 0.036 (range 0.815–0.904) against 0.792 for Vanilla Maxent, a mean improvement of +0.070. The advantage did not depend on a favourable seed: all ten runs exceeded the baseline, and the weakest of them still led it by +0.022, while the strongest led by +0.112. The full ROC curves show that the gain is distributed across the operating range rather than confined to one region as can be observed in figures \ref{fig:roc_auc_score} and \ref{fig:roc_auc_curves}.

\subsection{Classification at the 0.5 threshold}
\begin{figure*}[ht]
    \centering
    \includegraphics[width=1\linewidth]{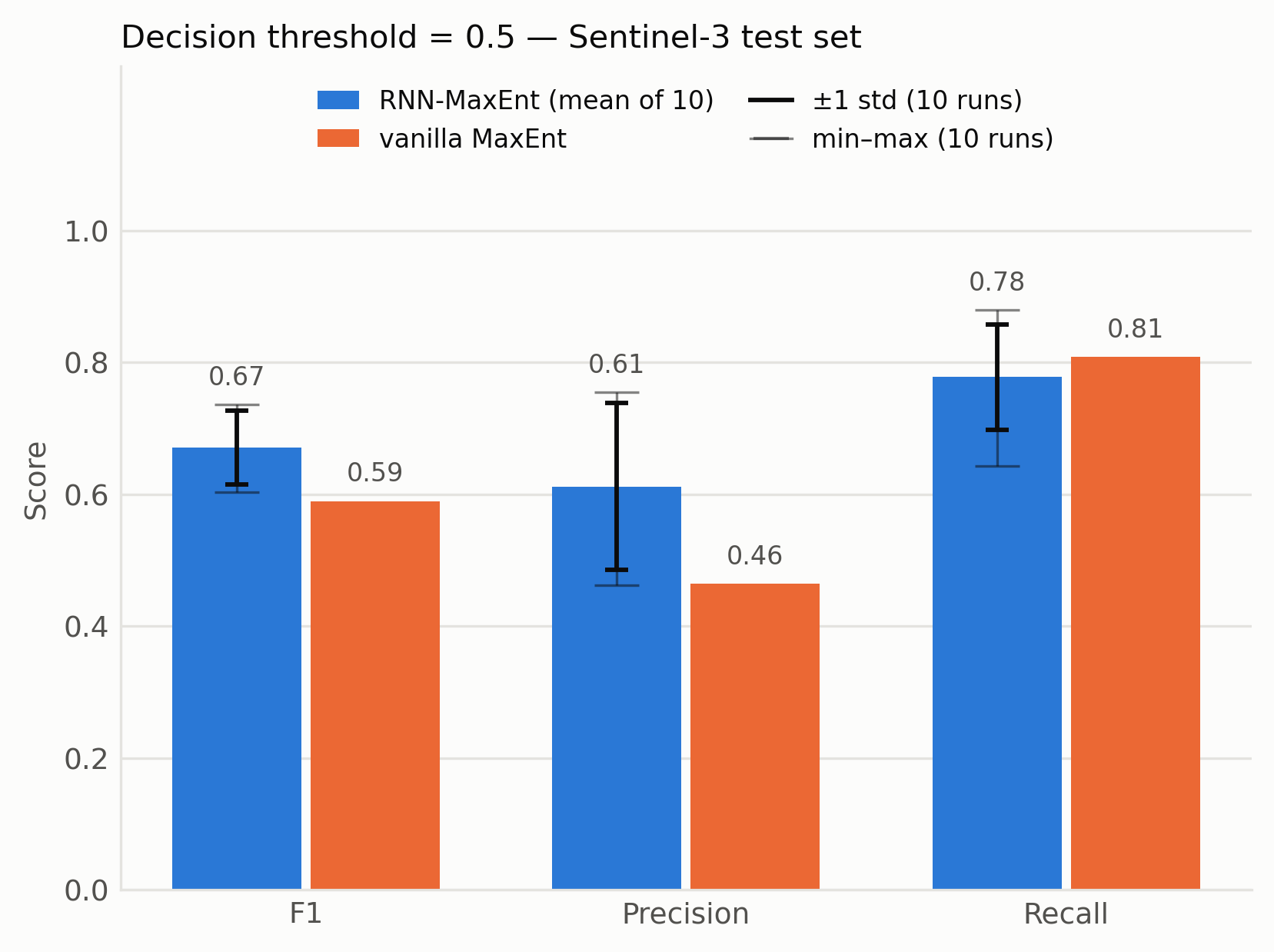}
    \caption{The figure shows the F1, precision and recall scores at a 0.5 decision threshold, RNN-Maxent's advantage is concentrated in precision rather than recall. F1, precision and recall for RNN-Maxent (blue, mean of 10 seeded runs) and Vanilla Maxent (orange, single fit) on the Sentinel-3 test set (n = 75,775). The thick whisker spans ±1 standard deviation across the 10 runs and the thin wide-capped whisker the observed minimum–maximum; numerals above each bar give the mean.}
    \label{fig:prf}
\end{figure*}

The output of both Vanilla Maxent and RNN-Maxent is confined in a range between 0 and 1, representing both the probability of DL presence and habitat suitability. We now report scores that consider every model output above 0.5 as DL presence, and every value below as DL absence. RNN–Maxent reached an F1 of 0.671 ± 0.056 (range 0.604–0.736) against 0.590 for the baseline, a mean gain of +0.081, with all ten runs above the baseline. The improvement is driven entirely by precision, which rose from 0.464 to 0.612 ± 0.126 (+0.148; 9 of 10 runs above the baseline). Recall was marginally lower, 0.778 ± 0.080 against 0.808 (−0.030), as shown in Figure \ref{fig:prf}.

The practical consequence is a large reduction in the number of samples flagged as suitable for the same detection rate. At 0.5 the vanilla baseline labels 51.1\% of all test samples as presence, while RNN-Maxent labels 39.6\% ± 12.5\%, and as little as 25.3\% in its most selective run. In absolute terms, the mean RNN–Maxent run retains 17,302 of 22,233 true presences while raising 12,716 false alarms, against 17,973 true presences and 20,727 false alarms for the baseline: 96\% of the baseline's sensitivity for 39\% fewer false alarms.

\section{Discussion on subsequent Analysis}

A recent review on the application of Earth observation to desert locust risk management highlights that EO-based studies have largely focused on habitat suitability assessment, while only a limited number of works have addressed broader questions such as socioeconomic loss estimation after infestations or vegetation damage analysis \cite{belfiore2025locust}{.}
At the same time, the review emphasizes the considerable potential of EO technologies for anticipatory action, including real-time situation assessment, targeted early response, outbreak forecasting, invasion prediction, and early warning support through the fusion of EO products with field-survey data. This perspective is consistent with earlier work on the desert locust upsurge in West Africa, which reviewed the early warning system developed for locust monitoring and discussed the conditions that allowed the outbreak to spread across such a large area \cite{ceccato2007locust}{.}
The study further assessed whether seasonal climate forecasting could extend the lead time for predicting locust development and movement, thereby improving preparedness and response. Although FAO-based early warning systems were already in place and had alerted the international community at the onset of the outbreak, the response remained too slow, allowing locust populations to increase rapidly and spread further. For this reason, locust management would benefit from either faster financing and implementation of control operations or enhanced predictive capability.\\
A further perspective shows how early warning systems have benefited from technological progress and improved field information over past decades. However, the same source also stresses that new limitations have emerged, particularly in recession areas where political instability, border disputes, insecurity, and conflict restrict access for survey and control teams. In many such areas, ground operations are either impossible or require military escort, which severely limits the ability to detect breeding or prevent invasions at an early stage. Moreover, the effectiveness of early warning depends on the presence of a well-organized and adequately funded national locust control structure capable of conducting surveys, rapidly transmitting geospatial data, using GIS for analysis, and maintaining timely communication with stakeholders through reliable information networks \cite{cressman2016desertlocust}{.}
In summary, sustainable improvements to locust early warning require analytical approaches that complement traditional survey-and-report-based systems. In this context, the integration of Earth observation products, climate data, and artificial intelligence offers a valuable alternative, because it can help overcome some of the logistical and security limitations affecting field-based monitoring while also improving transparency and reducing operational bias.
Subsequent analyses within early warning systems should leverage models such as the one presented in this study. Unlike traditional tools, the proposed RNN-Maxent model can provide historical, near-real-time, and forecast information on desert locust conditions across a continental domain in a form that is both accessible and operationally useful. More importantly, these outputs go beyond simple presence mapping: they can help experts interpret the climatic drivers behind specific locust behaviors and allow non-experts to identify emerging patterns in a more structured and intuitive way. This is the scope of the Desert Locust Monitoring Service  \cite{DLMSdestine}{,} which allows to forecast Desert Locust breeding hotspots in Africa and Asia. Building on this information, Desert Locust swarm migration is predicted thanks to an advanced stochastic model approach, described below.

\section{Migrations}
\begin{figure*}[ht]
    \centering
    \includegraphics[width=1\linewidth]{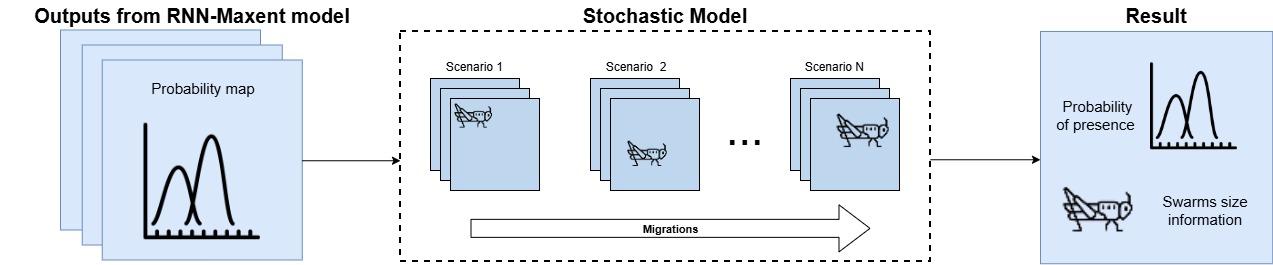}
    \caption{ On top of the RNN-Maxent model, a stochastic model is developed to simulate the physical and biological behavior of adult locusts and to predict potential migration patterns and swarm formation.}
    \label{fig:simple_locust_diagram_RNN-Maxent}
\end{figure*}

Desert locust swarms \footnote{The migration model is based on the gregarious locusts which are the locusts shaping swarms \cite{ symmons2001guidelines}{.} Therefore throughout the entire section it is not specified but implicit. } represent one of the most destructive migratory pest threats to global food security and agricultural economies \cite{worldbank2020locust}{.} A single swarm can travel up to 150 km per day and contain tens of millions of locusts, consuming vast quantities of vegetation and placing particular pressure on vulnerable farming systems in developing regions. Their impact is especially severe where environmental change may increase the frequency and spatial extent of favorable breeding and dispersal conditions \cite{worldbank2020locust}{.}\\
Climate change has further intensified concern about locust upsurges. Recent assessments indicate that unusually wet conditions, including rare cyclonic activity over the Arabian Peninsula and eastern Africa, contributed to vegetation growth that supported population multiplication and long-distance spread. In East Africa, anomalously high rainfall during 2019 helped facilitate rapid swarm expansion, while other swarms moved through Iran toward Pakistan and India. These developments have raised concern that warming oceans and more frequent tropical storms may increasingly create conditions favorable to outbreaks \cite{salih2020climate}{.}\\
Migration behavior in desert locust swarms has been documented for decades. Classic studies describe swarm flight activity and the environmental and biological conditions associated with long-range movement, providing an early conceptual basis for mechanistic models of dispersion \cite{kennedy1951migration}{.} Building on this foundation, physical and biological models have been developed to explain and simulate long-distance migration, including work on transport processes and flight dynamics \cite{lorenz2009migration}{.}\\
To move beyond traditional reporting, ground surveys, and isolated numerical approaches, more recent studies have increasingly combined remote sensing and geographically explicit monitoring with knowledge from remote sensing science, GIS, agronomy, plant safeguarding and meteorology \cite{chen2020geographic}{.} Technical references and monographs have further consolidated this interdisciplinary direction, highlighting the role of agricultural remote sensing and information technology in locust monitoring and analysis \cite{dong2023monitoring}{.} \\
In parallel, mathematical modeling has become increasingly important for forecasting swarm movement at short and long time scales. Recent integrated frameworks combine population dynamics, environmental and weather data, and atmospheric transport models to represent gregarious locust populations and their dispersion \cite{retkute2024framework}{.} These developments show that migration analysis can be formulated as a predictive problem supported by heterogeneous data sources and mechanistic assumptions. \\
Despite this progress, important limitations remain. Migration studies often rely on fragmented data sources, heterogeneous methods, and limited temporal coverage, which makes comparison across large regions difficult. Many high-resolution analyses are still restricted to regional domains, even though locust swarms can traverse very long distances in short periods. As a result, invasion events originating outside a study area may appear unexpected, even when they are consistent with broader migration dynamics. Understanding the spatial evolution of swarms is therefore essential not only for interpreting current movements, but also for anticipating subsequent upsurges.\\
To address these challenges, we propose a simplified stochastic model designed to integrate the information generated by the RNN-Maxent model (Figure \ref{fig:simple_locust_diagram_RNN-Maxent}). The aim is to generate coherent outputs and interpretable forecasts while retaining enough flexibility to represent the main drivers of migration. The results suggest that the RNN-Maxent suitable ecosystem model can complement traditional sources and support subsequent analyses.\\
The proposed model structure is described in the chapters below. For more details, including the physical assumptions and the datasets used, refer to Appendix B [\ref{sec:annex_migration}].

\subsection{Migration model concept: Biased Random  walk on a 2D lattice}
Modeling locust behavior is a challenge for statistical physics. Stochastic modeling on lattice can be used for simulating the locust outbreak based on the seasonal migration and the swarm aggregations. In this work, a  biased Random  walk on a 2D lattice  set-up to integrate with factual data is proposed \cite{kizaki1999stochastic}{.}\\
The geographical domain is represented as a two-dimensional square lattice, in which each lattice node corresponds to a pixel of the study area. This discretizations allows the spatial extent to be treated as a regular grid, where each cell can store the state of one or more locust groups. A locust group occupying a given pixel is represented by a node whose size is proportional to the group size, while the node color indicates the presence of locusts in that cell. If multiple groups occupy the same pixel, they are merged into a single aggregated group whose size reflects the total number of individuals.\\
Locust migration is modeled as movement across the lattice by successive discrete jumps. At each step, a group can move only to one of the eight neighboring cells, i.e. the Moore neighborhood in a regular grid. This representation captures local dispersion while preserving the spatial structure of the landscape. Movement is allowed only when the local conditions (described in \ref{sec:annex_migration}) at the current cell do not satisfy predefined stopping criteria, such as the maximum number of permitted jumps or environmental conditions that indicate settlement or termination of movement.\\
The direction of movement is stochastic but not purely random. Instead, it is modeled as a biased random walk, in which the probability of movement is influenced by wind direction. In this way, the model combines random local displacement with an externally driven directional preference, reflecting the role of atmospheric conditions in locust transport. The simulation continues until the defined number of steps is reached \footnote{Within a one time-step a group of locust can make multiple jumps.}. In this sense, the proposed model concept balances biological realism and computational simplicity, enabling parameterized dispersion simulations that integrate heterogeneous information and remain applicable over extensive spatial and temporal scales.

\subsection{Ingestion of early-stage RNN-Maxent model outputs}
The migration model is initialized using the daily outputs of the first RNN-Maxent model. These outputs are probability maps, where each pixel expresses the estimated likelihood that the corresponding location provides suitable conditions for early-stage locust development. To convert these probabilistic predictions into a spatial configuration on the lattice, each pixel is sampled stochastically: a locust group is assigned to a cell with a probability equal to the predicted suitability value for that pixel. This procedure generates a binary presence map for each day, where each cell is classified as occupied or unoccupied.\\
Because the migration model is designed to represent adult locust movement, the daily presence maps are aggregated over the developmental period required for locusts to reach adulthood [\ref{sec:annex_migration}]. During this interval, only days with favorable environmental and climatic conditions are retained, while it is assumed that locusts do not yet undertake long-distance migration, both because they have not reached the adult stage and because the lattice resolution is sufficiently coarse to neglect short-range displacement at this phase. The result is an initial configuration of adult locust groups distributed across the lattice, with each group assigned a standard size for the simulation.\\
This procedure is repeated periodically, according to the developmental time required for new adults to form bands. At each update interval, newly emerging adults are appended to the current lattice state before migration is simulated further. In practice, the model continuously ingests the outputs of the RNN-Maxent model and adds newly predicted adults to the simulation at each update step. Each update is generated through the same stochastic sampling scheme, so that alternative plausible configurations can emerge from the same probabilistic input. In this way, the model produces multiple migration scenarios over time rather than a single static realization. This recurrent integration captures the stochastic nature of outbreak initiation and allows low-probability suitable pixels to contribute to potential invasion pathways. This is particularly important because locust outbreaks may originate from rare but environmentally favorable conditions that become significant once local populations aggregate.

\subsection{Reducing the scenarios to single output}
\begin{figure*}[ht]
    \centering
    \includegraphics[width=1\linewidth]{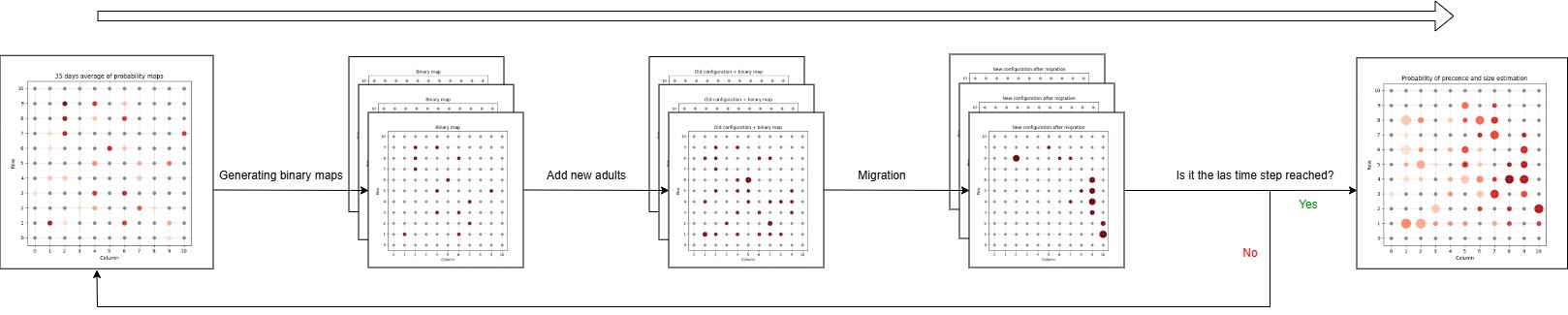}
    \caption{ From the RNN-Maxent model, configurations of newly emerged adults are continuously ingested into the migration model for each scenario, and the resulting migration patterns are then combined through statistical analysis to generate a single output.}
    \label{fig:complete_lattice_diagram_locust}
\end{figure*}

After all simulation scenarios have been generated and the maximum number of steps has been reached, the resulting trajectories are summarized through a statistical post-processing step. For each pixel of the lattice, the output map reports the probability that locusts appear in at least one simulated scenario. In addition, it provides summary statistics on group size across scenarios, including the minimum, maximum, and average values observed for each pixel (Figure \ref{fig:complete_lattice_diagram_locust}).\\
The probability of occurrence in at least one scenario is useful because it reduces the risk of overlooking plausible movement configurations or alternative dispersion trajectories. In this sense, it captures the breadth of the simulated uncertainty and highlights areas that may become affected under at least one valid migration pathway. The size statistics complement this information by describing the range of possible swarm dimensions, thereby allowing the map to represent best-case, worst-case, and most likely outcomes. \\
Compared with a simple presence/absence map, this output therefore provides a richer representation of potential migration outcomes, by combining spatial probability with quantitative information on swarm size.

\section{Conclusions}

In this work we introduced Neural network based Maxent, which extends Maxent's fixed feature dictionary with a Neural network learned feature map while preserving the model's presence-only statistical foundations and probability calibration. This lets the model learn nonlinear or temporal patterns, that Maxent would discard.

On Desert Locust habitat mapping, RNN-Maxent outperformed Vanilla-Maxent at the $0.5$ decision threshold: ROC-AUC $0.862 \pm 0.036$ vs.\ $0.792$ (all $10/10$ seeds above the baseline), F1 $0.671 \pm 0.056$ vs.\ $0.590$ ($10/10$), and precision $0.612 \pm 0.126$ vs.\ $0.464$ ($9/10$), at the cost of slightly lower recall ($0.778 \pm 0.080$ vs.\ $0.808$). The precision gain ($+0.148$) is roughly five times the recall loss ($-0.030$), and the improvement reflects sharper presence/background discrimination rather than a more conservative operating point: ROC-AUC is threshold-independent and still favours RNN-Maxent by $+0.070$, and a merely stricter model would buy precision by giving up F1, whereas F1 rises by $+0.081$. The drawback comes with the shift from a convex problem to a non-convex one, with high variance in the results depending on parameter initialization, minibatch ordering, and any other stochastic element. That variance is structured rather than random: seeds slide along the precision/recall trade-off and separate into two regimes.

The RNN-Maxent model demonstrated that a relatively simple AI-based approach can serve as a robust baseline for generating further analyses. Building on its outputs, the proposed stochastic migration model combines the suitability predictions of the RNN-Maxent framework with biological and environmental data, rather than relying solely on difficult-to-collect field reports. In this way, it can support the prediction of swarm occurrence and the estimation of swarm size. Such AI-based simulations are particularly valuable for traditional early warning systems, which often face substantial challenges in collecting ground information over large, remote, and hard-to-access areas.

The environmental suitability of desert locusts and their migration movements are mapped in the Desert Locust Monitoring Service which is accessible through the DestinE Platform at \url{https://dlms.sistema.destine.eu/}  \cite{DLMSdestine}{.}

\section*{ACKNOWLEDGMENTS}
We thank the FAO Desert Locust Information Service and the Locust Watch programme
for maintaining and making publicly available the Desert Locust observation records
used in this study.

We thank the International Centre of Insect Physiology and Ecology (\textit{ICIPE})
for their valuable feedback on the development of the model, whose expertise in desert
locust biology and monitoring helped guide and refine our approach.

The authors would also like to thank the "DestinE - Advanced Applications and Services Onboarding" program for supporting the technical implementation and maintenance of the Desert Locust Monitoring Service.

Destination Earth (DestinE) is a European Union funded initiative implemented by ESA, ECMWF and EUMETSAT. Access Destination Earth at \url{https://platform.destine.eu/}{.}

\clearpage
\printbibliography

\newpage
\section{Appendix A}

\subsection{Identification of Habitual Desert Locust Breeding Areas}
\label{sec:breeding_sites}
Only defining the suitable ecosystem can potentially lead to the identification of very large areas, just due to their climatic conditions, without taking into account historical likelihood of Desert Locusts presence. To narrow down the zones where Desert Locusts might be found, habitual breeding areas were identified by estimating the spatial density of historical desert locust early-stage observations using a kernel density estimation (KDE) framework applied independently within ecologically coherent
macroregions. For this step the full FAO Hoppers and Bands datasets were used, with the deduplication step applied, covering a period from 1985/02/25 to 2021/12/31.

\subsubsection{Macroregion Partitioning}
A well-known source of bias in the FAO database is the large variation in
survey intensity across countries: regions with dense monitoring networks
contribute disproportionately many records, which in a global KDE would
suppress the apparent density of less-monitored areas. To mitigate this
effect, the study domain ($26\degree$W--$76\degree$E, $4\degree$S--$38\degree$N)
was partitioned into seven ecologically and operationally coherent macroregions: West Africa, North Africa, Nile \& Sudan, Horn of Africa, Arabian Peninsula, Iran, and South \& Central Asia. Region boundaries were derived from the Natural Earth 50\,m administrative dataset \cite{naturalearth}{.} Each observation was assigned to the macroregion containing its (snapped) coordinate using a point-in-polygon test; the small fraction of records falling outside all defined regions was discarded.

\subsubsection{Kernel Density Estimation}
A Gaussian KDE was fitted separately for hoppers and bands within each macroregion. To ensure a physically meaningful, isotropic bandwidth selection, all observation coordinates were first projected from geographic (WGS\,84 longitude--latitude) to a Lambert Azimuthal Equal-Area (LAEA) coordinate system centred at $25\degree$E, $17\degree$N---the approximate centroid of the study domain---using the \texttt{pyproj} library. In this projection both axes are in metres, so Scott's rule~\cite{scott1992density} for automatic bandwidth selection operates on distances that are comparable across the two spatial dimensions. The evaluation grid, defined at $0.1\degree$ resolution in geographic coordinates, was likewise projected to LAEA before evaluation, and the resulting density values were
mapped back to the regular geographic grid for display and further analysis.

Macroregions with fewer than 30 observations of a given type (hoppers or bands) were excluded from the corresponding KDE to avoid bandwidth under-determination. All seven macroregions exceeded this threshold for both record types.

\subsubsection{Hotspot Delineation}
Within each macroregion the KDE surface for hoppers and the KDE surface for bands were combined and normalized to the interval $[0, 1]$ by a min--max transformation. The two normalised surfaces were then combined as a sum.

The seven regional surfaces were then merged into a single global map by taking, at each grid cell $\mathbf{x}$, the maximum value across all the covered regions:

\begin{equation}
    D(\mathbf{x}) = \mathrm{norm}_{[0,1]}\!\left(
        \max_{r}\, D_r(\mathbf{x})
    \right).
\end{equation}

\begin{figure*}[ht]
    \centering
    \includegraphics[width=1\linewidth]{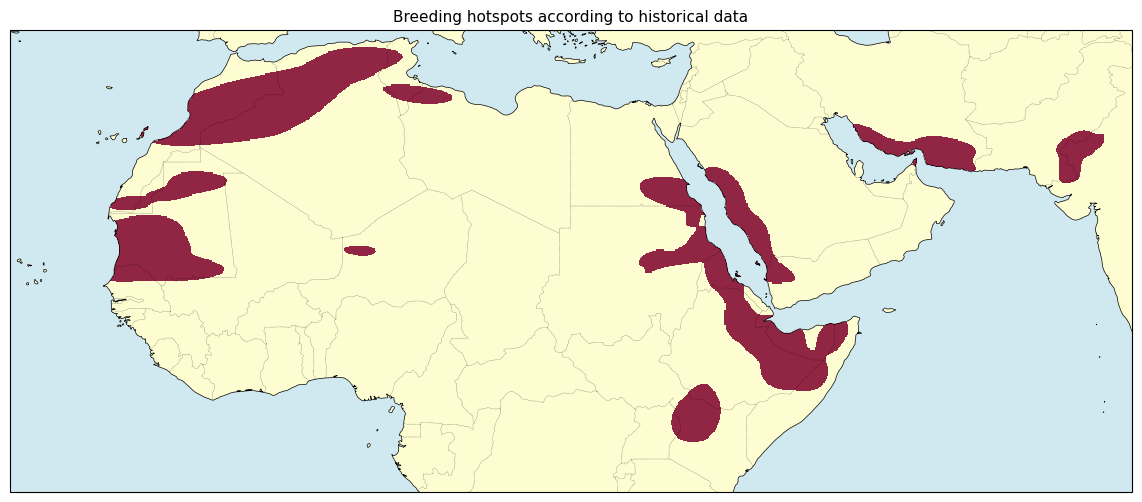}
    \caption{Areas where, according to historical data from 1985/02/25 to 2021/12/31, Desert Locusts have been most often found.}
    \label{fig:breeding_hotspots}
\end{figure*}

Grid cells with a normalised density exceeding a heuristically defined threshold were classified as habitual breeding hotspots. Ocean and inland water cells were masked using a land polygon derived from the Natural Earth 50\,m physical dataset. The resulting binary mask delineates the areas that concentrate the historically observed breeding activity, shown in Figure \ref{fig:breeding_hotspots}.

\newpage
\section{Appendix B}

\subsection{Assumptions}
\label{sec:annex_migration}

The assumptions adopted in this study are based on gregarious desert locust swarms \cite{symmons2001guidelines} and related FAO reference material, and were selected conservatively with a focus on worst-case conditions because the primary objective of the model is early warning and outbreak prevention. For this reason, locust mobility and outbreak potential were intentionally biased toward higher-risk configurations, and several parameter values were rounded upward where appropriate. During the testing phase, selected parameters were further adjusted to improve consistency with FAO bulletins and reported locust dynamics. More generally, the model was designed to remain as simple as possible while still capturing the main drivers of locust migration, so that it can serve as an operational baseline and later be refined with domain experts. In this way, the core dynamical structure is preserved while thresholds, conditions, and parameter values can be adapted in future versions. For instance, the framework was also implemented at different temporal resolutions, including daily, weekly, and 10-day time steps, by aggregating input data and maps accordingly. This makes it possible to evaluate the trade-off between temporal precision and computational efficiency while keeping the model flexible enough for continental-scale applications.
A locust group is assumed to leave a pixel only when the local conditions are favorable for dispersion. Specifically, movement is allowed when food availability is insufficient, wind conditions are strong enough to support take-off, and temperature is suitable for flight.  Conversely, a group remains in the same pixel if food resources are sufficient, if wind or temperature conditions are not favorable for dispersion, or if the group has already reached its maximum possible travel distance. These conditions are represented in the model by the idea that movement is driven by a conjunction of environmental and physiological triggers.
\begin{align*}
&\text{Food availability is sufficient if Leaf Area Index (LAI)} > 0.065 \\
&\text{Wind speed is strong enough if } \text{speed\_wind} > 10 \text{ m/s} \\
&\text{Temperature at 850 hPa is suitable if } 283.15\,K < \text{Temperature} < 313.15\,K \\
&\text{Maximum daily traveled distance if } \text{traveled\_distance} < 100 \text{ km/day}
\end{align*}

Each simulation incorporates all the outputs generated during the previous 5 months, as this is the amount of time assumed to be the lifespan of an adult locust.
In addition, the time required for an early-stage locust to become an adult is assumed to be $\delta35$ days. This period defines the update interval at which newly matured adults are ingested from the first model and introduced into the migration simulation. These assumptions allow the model to link the emergence of new adults with the timing of dispersion.\\
Finally, each locust group in the initial configuration is assigned the same size, corresponding to an average band size of 10 ha. When groups merge, their sizes are summed, and the resulting aggregated group is considered a swarm once it exceeds the threshold of 50 ha. This threshold-based representation captures the transition from smaller bands to larger migratory units while keeping the simulation computationally tractable.

\subsection{Calibration}
Reports of desert locust on ground are needed for calibrating the parameters, thresholds and assumptions described so far. In our work we propose a calibration methodology based on the FAO Desert Locust Information Service, that shall be extended to other reports for overcoming the strict dependency on the quality and biases of the FAO data.\\
During the calibration phase, the model was seeded from FAO records during a 10-day window and run forward daily for 30 days. Predicted locust positions were then scored against subsequent FAO observations at T+10, T+20, and T+30 days using hit rate metrics at four distance tolerances (50, 100, 200, and 400 km) and median nearest-prediction distance. This process was repeated across three consecutive calibration windows per period, giving nine scoring points per parameter combination and producing a robust ranking of parameter sensitivity.\\
The calibration identified wind deviation probability and the LAI stopping threshold as the two most influential parameters. A wind deviation of 0.5 consistently outperformed lower values across both validation periods, confirming that locusts do not follow a single wind vector rigidly and that stochastic directional spread is necessary. The LAI stopping threshold of 0.065 was consistently optimal; increasing it to 0.300 alone caused a 30–40 percentage point drop in hit rate at 200 km, confirming that the model correctly represents the sparse semi-arid vegetation conditions under which desert locusts migrate. To assess the final model performance, calibration was performed on October–December 2025 data (West Africa outbreak) and independently validated on January–March 2025 data (East Africa and Arabian Peninsula outbreak) which represents a geographically distinct region and season, with no overlaps. The calibration period achieved a best hit rate of 91.8 $\%$ at 200 km with a median prediction distance of 32.8 km. On the independent validation period, the model achieved 78.1 $\%$ at 200 km with a median distance of 65.4 km, confirming genuine predictive skill across regions.

\subsection{Data Pool Migration Model}
The intrinsic complexity of the model made data preparation and integration a central part of the work. Because the objective was to produce forecasts whose quality depends strongly on antecedent seasonal conditions, the model combines historical, near-real-time, and forecast data within a single workflow. A further challenge was the heterogeneity of the input sources, which required harmonization before they could be jointly used. In particular, all datasets were resampled or aggregated to a spatial resolution consistent with the first model, namely 5.5km. This ensured that the outputs of the suitability model and the inputs to the migration model could be combined in a coherent way.\\
Data gaps also had to be addressed, especially for variables such as vegetation, where missing observations can affect downstream modeling. To mitigate this issue, common aggregation and gap-handling techniques (e.g., temporal interpolation) were applied before the data were ingested into the simulation framework. In addition, one of the key components of the workflow was the use of Destination Earth data, which provided essential information for the construction of the model inputs. Finally, Table \ref{tab:data_providers} shows the data used in the migration model.

\begin{table}[h]
\centering
\caption{Data providers, datasets, and variables used in the migration model.}
\label{tab:data_providers}
\begin{tabular}{lll}
\toprule
\textbf{Data Provider} & \textbf{Dataset} & \textbf{Variable} \\
\midrule
DestinE & CCADT & Temperature \\
DestinE & CCADT & U wind component \\
DestinE & CCADT & V wind component \\
DestinE / Copernicus & EO.CLMS.DAT.GLO.LAI300 & Leaf Area Index (LAI) \\
DestinE & Layer 1 Output & Early-Stage Locust probability \\
\bottomrule
\end{tabular}
\end{table}

\end{document}